\documentclass[lettersize]{article}
\UseRawInputEncoding
\pdfoutput=1
\usepackage{amsmath,amsfonts}
\usepackage{algorithmic}
\usepackage{array}
\usepackage[caption=false]{subfig}
\usepackage{textcomp}
\usepackage{stfloats}
\usepackage{url}
\usepackage{verbatim}
\usepackage{graphicx}
\usepackage{caption}
\usepackage{authblk}
\usepackage{lineno}
\usepackage[misc]{ifsym}

\usepackage{booktabs}
\usepackage{color}
\usepackage{hyperref}
\usepackage{listings}
\usepackage{balance}

\begin{document}
	\title{Next-Generation Simulation Illuminates \\Scientific Problems of Organised Complexity}
	\author[1,2,3 ${\textrm{\Letter}}$]{Cheng Wang}
	\author[1]{Chuwen Wang}
	\author[1]{Wang Zhang}
	\author[1]{Yu Zhao}
	\author[1]{Shirong Zeng}
	\author[1]{Ronghui Ning}
	\author[1,2,3]{Changjun Jiang}
	
	\affil[1]{Department of Computer Science and Technology, Tongji University, Shanghai, China}
	\affil[2]{Key Laboratory of Embedded System and Service Computing, Ministry of Education, Shanghai, China}
	\affil[3]{Shanghai Artificial Intelligence Laboratory, Shanghai, China}
	\affil[${_\textrm{\Letter}}$]{e-mail: cwang@tongji.edu.cn}
	
	%\markboth{Journal of \LaTeX\ Class Files,~Vol.~18, No.~9, September~2020}%
	%{How to Use the IEEEtran \LaTeX \ Templates}
	\date{}
	
	\maketitle
	
	\begin{abstract}
		 As artificial intelligence becomes increasingly prevalent in scientific research, data-driven methodologies appear to overshadow traditional approaches in resolving scientific problems. In this Perspective, we revisit a classic classification of scientific problems and acknowledge that a series of unresolved problems remain. Throughout the history of researching scientific problems, scientists have continuously formed new paradigms facilitated by advances in data, algorithms, and computational power. To better tackle unresolved problems, especially those of organised complexity, a novel paradigm is necessitated. While recognising that the strengths of new paradigms have expanded the scope of resolvable scientific problems, we aware that the continued advancement of data, algorithms, and computational power alone is hardly to bring a new paradigm. We posit that the integration of paradigms, which capitalises on the strengths of each, represents a promising approach. Specifically, we focus on next-generation simulation (NGS), which can serve as a platform to integrate methods from different paradigms. We propose a methodology, sophisticated behavioural simulation (SBS), to realise it. SBS represents a higher level of paradigms integration based on foundational models to simulate complex systems, such as social systems involving sophisticated human strategies and behaviours. NGS extends beyond the capabilities of traditional mathematical modelling simulations and agent-based modelling simulations, and therefore, positions itself as a potential solution to problems of organised complexity in complex systems. 
	\end{abstract}
	
	\section{Main}
	
	As Warren Weaver elucidated in ``Science and Complexity'' \cite{weaver1948science}, scientific problems can be classified into three categories: problems of organised simplicity, problems of disorganised complexity, and problems of organised complexity, each distinguished by their unique levels of intrinsic complexity and randomness. The quest for resolving more problems has led to the continual proposal of new methodologies, which is a destined result of the comprehensive progress in data, algorithms, and computational power.
	
	Data, algorithms, and computational power are the cornerstones of scientific development, and these three elements are closely related to \textit{the four paradigms} \cite{hey2009fourth}. Based on the observation of natural phenomena and experimental summary, the integration of data and algorithms formed the first paradigm---empirical science \cite{thibault2016self}. With the improvement of scientists' ability in algorithmic design, the integration of algorithms and computational power to explain the principles behind certain natural laws formed the second paradigm---model-based theoretical science. Subsequently, the algorithm development and increase in computational power gave rise to the third paradigm---computational science \cite{gahegan2020fourth}. Finally, with the further development of computers, the ability to collect, store, and process large amounts of data has dramatically changed science. A high degree of integration of data, algorithms, and computational power achieved the fourth paradigm---data-driven science \cite{agrawal2016perspective}.
	
	Subsequent to the publication of Weaver's work in 1948, the advent of the third and fourth paradigms has yielded significant advancements in tackling problems of organised complexity. Methods of either paradigm partially possess the ability to analyse the uncertainty beyond completely randomness which follows some specific distribution instead of typical distributions like normal distribution. However, the comprehensive development of the three elements indicates the stagnation of paradigms \cite{zhang2023scaling}. To better research problems concerning higher levels of uncertainty, the introduction of new mechanism is imperative. We suggest integration as such a mechanism, enabling the exploration of these problems through collaborative efforts of multiple paradigms to compensate for each other's limitations. When targeting problems of organised complexity within more complex systems, the combination of simulation and machine learning stands out as a promising approach \cite{haasl2022simulation}, providing an efficacious platform to actualise the concept of paradigms integration and empowering more authentic simulation to better analyse complex systems. This simulation of paradigms integration can be the \textit{next-generation simulation} (NGS). It enables the agents in the simulated system to have dynamic behaviour patterns learnt during the simulation to better represent the uncertainty in complex system than existent simulation technologies where agents' behaviour are of specific rules or fixed distributions. 
	
	Inspired by the explosion of foundation models and LLM-based agent simulation \cite{2021opportunities, park2023generative}, we propose \textit{sophisticated behavioural simulation} (SBS) as a methodology to realise the NGS. Within SBS, ``sophisticated'' refers to complex experience-related features of human behaviours. It is an integration of the second paradigm, the third paradigm, and the fourth paradigm. Within SBS, methods of the second paradigm formulate the fundamental rules of a system; methods of the fourth paradigm realise authentic individual behaviours in systems through our proposed model tower. Methods of the third paradigm combine these two elements making heterogeneous simulations. We believe that SBS will be unprecedentedly powerful in analysing problems of organised complexity in complex human systems that are always of high-level uncertainty. 
	
	In this Perspective, we rethink the existent scientific paradigms through the lens of data, algorithms, and computational power. We discuss the challenges of addressing problems of organised complexity from a standpoint of uncertainty. Through pertinent examples, we think that the paradigms integration is a possible solution to problems of organised complexity, and the NGS is one promising form of paradigms integration. Accordingly, we propose SBS to realise the NGS and demonstrate the implementation and utilisation methods of SBS. 
	
	\section{Organised Complexity}

	\begin{figure}[!t]
		\centering
		\includegraphics[width=0.75\columnwidth]{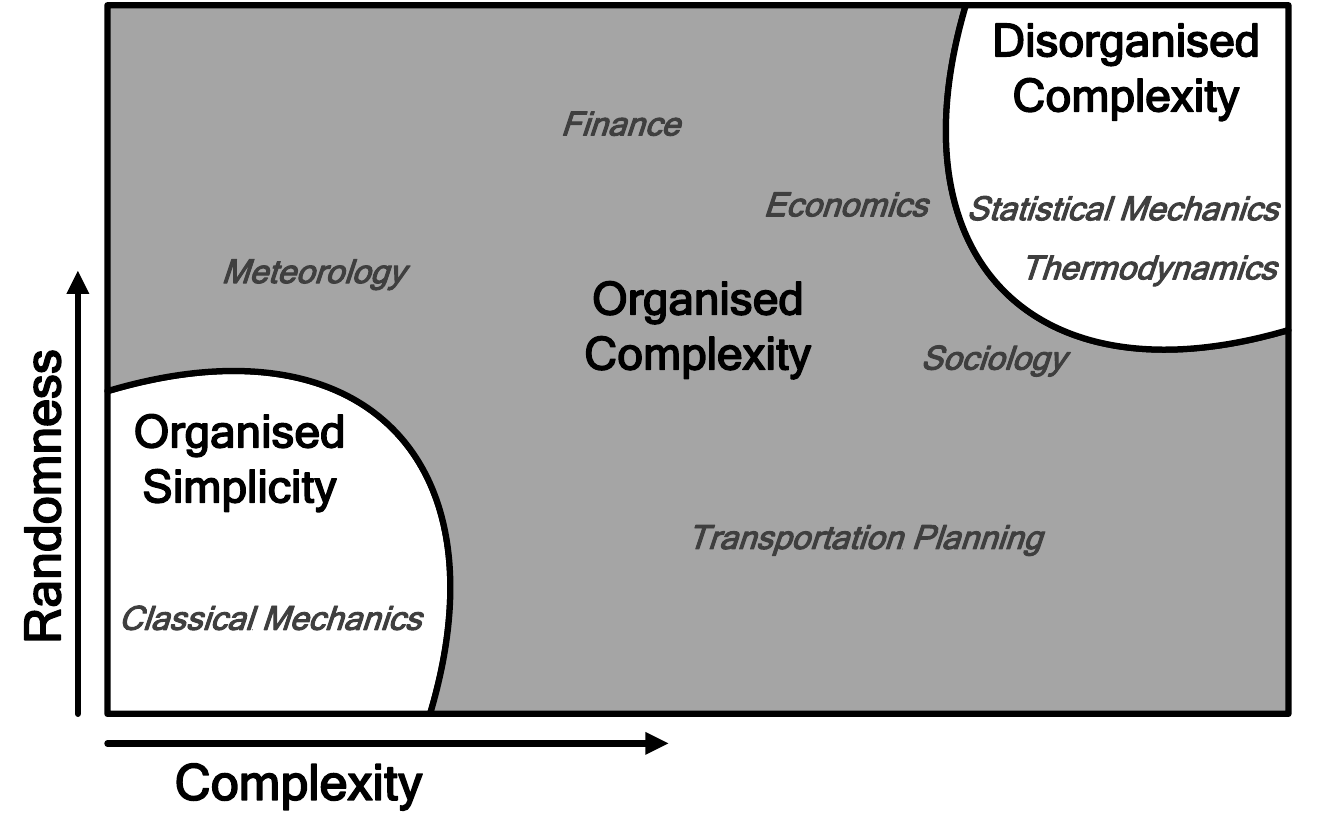}
		\caption{Three classes of systems and their examples.}
		\label{fig:OC}
	\end{figure}
	
	Throughout the evolution of science, complexity and randomness have emerged as two pivotal criteria for characterising the nature of scientific problems \cite{weaver1948science}. Complexity refers to the number of variables implicated in a problem, while randomness refers to the unpredictability or disorderliness of each variable within a system. The classification based on these criteria offers a clear and robust framework for comprehending the essence of scientific problems. The correlations between complexity, randomness, and systems are depicted in Figure \ref{fig:OC}. 
		
	The problems of organised simplicity and the problems of disorganised complexity represent two extremes, with a category known as the problems of organised complexity existing between them \cite{klir2006uncertainty}. Problems of organised simplicity can be analysed with formulas and equations thanks to the limited variables and simple relationship \cite{newton1833philosophiae}. Problems of disorganised complexity can be analysed with statistical methodologies because large quantities of variables exhibits stable features in such systems \cite{pathria2016statistical}. The number of variables involved, although fewer than that of problems of disorganised complexity, far exceeds that of problems of organised simplicity. In contrast to disorganised problems, each variable and its interactions in problems of organised complexity exhibit a certain degree of organisation and regularity. This implies that, despite the numerous factors within the system, the relationships between these factors are not completely random but possess a certain order and predictability.
	
	In reality, problems of organised simplicity and problems of disorganised complexity represent only a small fraction of scientific problems. Many problems in human society fall into the category of organised complexity, with economics being a particularly prominent example. Considering the seemingly straightforward issue of agricultural product pricing, this involves the influences of various factors such as policies, production levels, and the market, with these factors interconnecting in a complex and organic manner. At the theoretical level, Hausmann et al. \cite{hidalgo2009building} introduced two fundamental concepts of economic complexity: relatedness and complexity. Measurement of economic complexity not simply involves adopting aggregation (corresponding to simple systems) or distribution (corresponding to complex systems). It involves the use of dimensionality reduction techniques to preserve the characteristics of the involved elements and consider their interactions.
	
	Besides economics, there are also problems of organised complexity in sociology, transportation, etc. In sociology, intricate elements such as interactions among social actors and cultural exchanges form complex social networks \cite{freeman2004development,emirbayer1994network,vega2007complex}. In transportation, factors such as traffic flow and road planning interact, leading to the emergence of complex characteristics within the transportation system \cite{chen2012comprehensive,tang2013characterizing}. 
	
	These instances underscore the pervasive presence of organised complexity within complex human systems. Addressing these complexities necessitates a comprehensive consideration of numerous factors and a profound understanding of the interdependencies among these factors, thereby posing new challenges to scientists. Warren Weaver asserted that resolving such problems demands scientific advancements surpassing those necessitated by the other two categories of problems. The subsequent evolution of scientific paradigms confirmed his assertion.
	
	\section{The Evolution of Scientific Paradigms}
	
	\begin{figure}
		\centering
		\includegraphics[width=1.0\columnwidth]{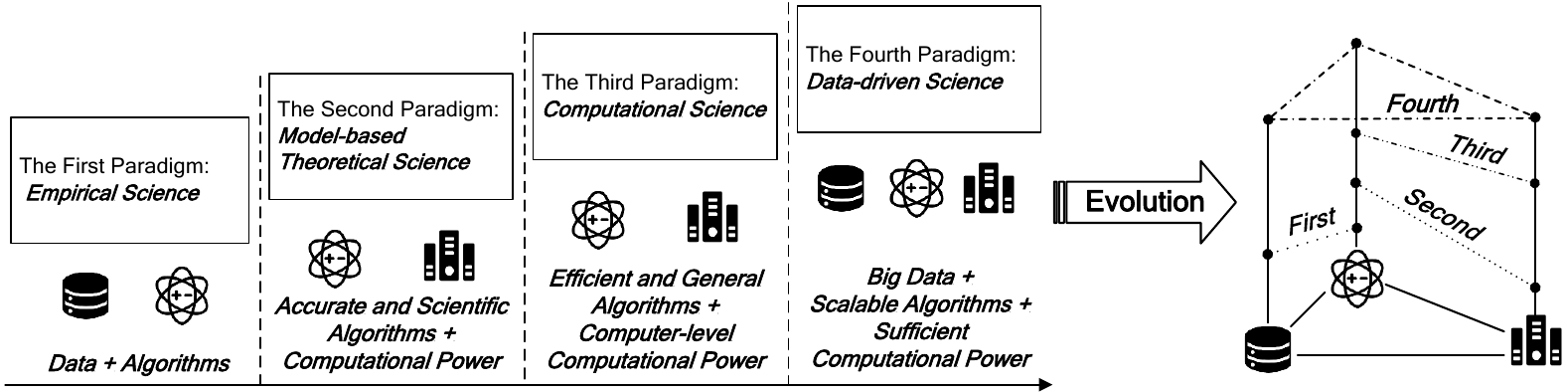}
		\caption{The evolution of the four paradigms aligned with the three elements.}
		\label{fig:paradigm}
	\end{figure}
		
	The concept of scientific paradigm was originally introduced by the renowned American philosopher of science, Thomas Samuel Kuhn \cite{kuhn2012structure}, in 1962. Paradigm refers to the theoretical foundation and practical norms upon which conventional science operates, constituting the world view and behavioural patterns collectively adhered to by a community of scientists. Turing Award laureate Jim Gray proposed four scientific paradigms in his final lecture and used them to describe the historical evolution of scientific discovery \cite{hey2009fourth}. Four paradigms are empirical science, model-based theoretical science, computational science, and data-driven science. Figure \ref{fig:paradigm} illustrates the evolution of the four paradigms aligned with the development of three elements---data, algorithms, and computational power. The four paradigms, each adopting distinct perspectives and methodologies, have facilitated addressing a wide range of scientific problems.
	
	The first paradigm, termed empirical science \cite{thibault2016self}, is fundamentally empirical, characterised by the recording and description of natural phenomena. It hinges upon direct observation and hands-on experimentation within the natural world. Although numerous apparent rules are observed, there lacks systematic methods to capture or express them. However, the disadvantage in all three elements makes it inefficient and challenging to achieve a more precise understanding of phenomena.
	
	With the improvement of scientists' capabilities of algorithm, induction, and deduction, the second paradigm, known as model-based theoretical science, emerged. In comparison with the first paradigm, which relies on observation and experimentation to discern rules, the scientific theories of the second paradigm require principled explanations behind certain natural rules. Examples include Newton's laws of motion from the 17th century \cite{newton1999principia} or Maxwell's electrodynamics equations from the 19th century \cite{maxwell1865viii}. These theoretical models are inductively derived from empirical observations and can be generalised to broader situations to deduce more theory. However, when it comes to problems of organised complexity where complex interactions cannot be ignored, the difficulty of validating theories increases, diminishing the effectiveness of the second paradigm.
	
	In the 20th century, the rapid development of computers brought about a significant increase in computational power and algorithm capabilities, giving rise to the third paradigm---computational science \cite{gahegan2020fourth}. Around the mid-20th century, John von Neumann proposed modern electronic computer architecture, which resulted in the widespread adoption of simulation and emulation for scientific experiments using electronic computers \cite{von1993first}. As computer simulation increasingly replaced experiments, the third paradigm gradually became a conventional method in scientific research. Methodologies of the third paradigm are a better solution to some problems of organised complexity in natural science. Numerical analysis realised through mathematical models and enough computational power eliminates the bottleneck of dealing with large amounts of complex interactions in a system. However, mathematical models are insufficient to simulate all the patterns of the real world.
	
	In the early 21st century, the development of computers once again transformed science. The ability of computers to collect, store, and process large amounts of data gave rise to the data-intensive fourth paradigm---data-driven science \cite{agrawal2016perspective}. Methodologies of the fourth paradigm further resolve some problems of organised complexity through pattern recognition, such as image classification and temporal prediction. The complex interactions in a specific problem are represented as a complicated distribution.
	
	The four scientific paradigms reveal an ongoing evolutionary trajectory in scientific research. Each paradigm holds a significant position within its respective era, expanding the scope of resolvable problems. Although they progressively reduce the area of unresolved problems, the comprehensive advancement of the three elements indicates a limited progression in addressing further problems of organised complexity. The scaling of data, algorithms, and computational power can only yield more powerful methodologies within the fourth paradigm \cite{zhang2023scaling}, thereby failing to fundamentally resolve problems that cannot be transformed into specific distribution-finding tasks on big data. The four paradigms are now facing this bottleneck.
	
	\section{Bottleneck Arising From Uncertainty}
	
	Uncertainty refers to the unpredictability of individual and systemic states. It is a ubiquitous characteristic in systems of organised complexity or disorganised complexity. Not all problems of uncertainty are intractable. For example, in Thermodynamics, temperature is a physical quantity that quantitatively expresses the characteristics of random motions of microscopic particles. Its effectiveness results from the fact that random motions consistently follow a normal distribution, and these motions can be regarded as a whole when scaled up. Similarly, it is the relatively fixed distribution inherent in the uncertainty that makes the methods of the fourth paradigm, data-driven science, effective for tasks like image classification and generation.
	
	However, in most systems of organised complexity, there is no fixed distribution. The absence of a fixed distribution leads to higher levels of uncertainty. The theory of chaos serves as a pertinent demonstration. In 1962, Lorenz \cite{lorenz1963deterministic} observed a phenomenon where a finite deterministic system, representing forced dissipative hydrodynamic flow through ordinary non-linear differential equations, exhibited instability. This instability manifested as non-periodic solutions that diverged significantly from bounded solutions with minor differences in initial states. Subsequently, May \cite{may1974biological} explored systems described by different equations and identified similar instabilities while studying biological populations. Two years later, May \cite{may1976simple} further demonstrated the potential for creating non-deterministic systems even with simpler mathematical models. In essence, chaos is the union of the transience of stability, the sensitivity of initial states, the general existence of non-periodic state sequences and the very-long-range unpredictability of non-idealised systems. Consequently, complex systems of this property are called chaotic systems. While some problems of this level of uncertainty have been partially resolved, such as improved weather forecasting in recent decades through the advancements in simulation and deep learning, bottleneck remains.
	
	After demonstrating the concept of chaos, we can explore its extension, high-order chaos. Different from individual simulacra of specific expression or fixed distribution in systems such as Meteorology, human behaviours are notably intricate. Research conducted in 1990 \cite{richards1990strategic} revealed that human decision-making within classical game theory systems is heavily influenced by others' behaviours, often deviating from the anticipated outcomes in Nash equilibrium scenarios \cite{nash1951non}. This indicates human behaviours are dynamic in complex decision-making scenarios. Existent methods of the third paradigm or the fourth paradigm primarily rely on theories derived from induction, idealised hypotheses, and fixed distribution of behaviours \cite{yuan2023generating, 398900, kang2014biocellion, hui2022knowledge, yuan2022activity}, limiting their capacity to authentically simulate human behaviours in these scenarios. When applied to complex systems such as stock markets and social networks, they may suddenly encounter failures due to the oversimplification of human behavioural process.
	
	The higher level of uncertainty within human-strategy-related systems can be regarded as a form of high-order chaos, where every individual within the chaotic system is itself chaotic. There exists a bottleneck that the four paradigms are not capable of resolving problems concerning high-order chaos.

	\section{Integration of Scientific Paradigms: A Paradigm Beyond Four Existent Paradigms}
	
	Confronted with the bottleneck arising from high-level uncertainty, existing methods of the second paradigm, the third paradigm, and the fourth paradigm prove inadequate. In these scenarios, which include fields like sociology and economics, researchers have to resort to the first paradigm that emphasises observing the real world and inducing laws. The limitations of the first paradigm are evident. Real-world experiments are often limited, inefficient, and costly. Since the three elements have well developed, we suggest the integration of the four paradigms as a new paradigm to eliminate the bottleneck.
	
	With the scientific problems under research having become more complex. Researchers employing research methods of different paradigms often find that a single method cannot fully address these challenges. This necessitates collaborative efforts, integrating research methods of different paradigms to capitalise on their respective strengths and compensate for limitations, which demonstrates the advantages of complementation among scientific paradigms. The deciphering of the deoxyribonucleic acid (DNA) structure marks a significant milestone in the history of biology. Its discovery process involves exploration through research methods of different paradigms. Scientists represented by Rosalind Franklin and Maurice Wilkins captured and observed the DNA structure adopting an X-ray diffractometer, i.e., the first paradigm research method through observation and analysis of experiments \cite{franklin1953molecular}. Scientists like James Watson and Francis Crick explored the DNA structure by deriving models based on chemical rules, utilising the second paradigm through model deduction. After extensive research on their methods, neither side fully revealed the truth. Eventually, collaboration ensued: Watson and Crick \cite{watson1953molecular} utilised reports from Wilkins and Franklin along with high-resolution DNA photos taken by Franklin to remodel and deduce the DNA double helix. Simultaneously, Franklin and Raymond Gosling \cite{cobb2023rosalind} provided crystallographic evidence supporting the double helix structure, while Wilkins further validated the model's reliability through DNA crystal experiment data. Besides, AlphaFold2 predicted the distance distribution between each pair of amino acids in a protein, along with the angles between the chemical bonds connecting them, and then compiled all the measured results of amino acid pairs into 2D distance histograms. Convolutional neural networks learnt from these images to build 3D structures of the protein \cite{jumper2021highly}. The combination of this first paradigm research method with the fourth paradigm research method effectively predicts protein structures. The combination of two paradigms, as evidenced by the three examples mentioned, offers a complementary approach that leverages strengths to compensate for weaknesses. This synergy proves essential.
	
	In conclusion, the four paradigm research methods each hold their unique value, and therefore, can be integrated and complemented. We strives to achieve a higher degree of integration among the four paradigms, thereby establishing a novel research methodology that leverages their collective strengths to resolve problems of organised complexity. Drawing upon the success of the third and fourth paradigms, we posit that a new simulation technology that harnesses the power of paradigms integration holds promise.
	
	\section{Next-Generation Simulation: A Possible Solution to Problems of Organised Complexity Based on Paradigms Integration}
	
	Before the invention of computers, computational efficiency was severely limited. The emergence of early computers shattered these constraints, bringing about computational speed and efficiency exceeding the limits of human capability. This significant transformation set the stage for the birth of simulation technologies. Within the Manhattan Project during World War II \cite{goldwhite1986manhattan}, simulation technology based on analogue computers heralded the implementation of the first generation of simulation technologies \cite{balachandra2000introduction}.
	
	The first generation of simulation primarily operated within idealised conditions, focusing on relatively simple systems. During the Manhattan Project, ENIAC utilised the Monte Carlo algorithm to simulate the explosion process of the atomic bomb, employing a random number generator to mimic stochastic elements like neutron transmission and fission reactions \cite{benov2016manhattan}. Similarly, simulation conducted on analogue computers contributes to the research of neutron paths \cite{coccetti2016fermiac}, non-linear mechanical vibrating systems \cite{5059991}, aircraft engine performance \cite{ketchum1952simulation}, nuclear reactor start-up \cite{4315572}, etc.
	
	However, the constraints in computational speed and storage capacity restrict the first generation of simulation to relatively simple physical models. For complex systems and phenomena, such as fluid dynamics, large-scale structural mechanics, etc., it was difficult to carry out precise simulations. The emergence of the second generation of simulation greatly alleviated this problem.
	
	The second generation of simulation greatly benefited from the development of supercomputers. After the end of World War II, the computer industry entered a phase of rapid development, which led to significant advances in computer storage capacity and computational power. Take the example of computational power, measured by floating point operations per second (FLOPS), which advanced from the age of KFLOPs \cite{eckert1959design, lukoff1959design} in the 1950s to GFLOPs \cite{august1989cray} in the 1980s. Supercomputers and normal computers have essentially the same components but differ greatly in scale. The main characteristics of supercomputers include two aspects: great data storage capacity and extremely fast data processing speed \cite{segall2015research}. It is these characteristics that enable supercomputers to simulate a complex system that comprises huge amounts of entities.
	
	The application scope of the second generation of simulation technology is remarkably broad. It spans from simulating climate systems \cite{398900} and earthquake disasters \cite{5644908} to biological systems \cite{kang2014biocellion} and chemical systems \cite{10046049}. Supercomputers extend the applicability of simulation to non-linear dynamical systems without closed-form expressions, relying exclusively on numerical analysis. This pivotal advancement enables researchers to reliably characterise the evolution of complex systems through simulation, facilitating a dependable assessment of the states of complex systems, and this is exactly the third scientific paradigm---computational science. The effectiveness of the third paradigm also promotes the spread and development of simulation in turn. Scientists no longer have to rely entirely on empirical or theoretical research but can use direct observation and mathematical modelling to visualise and simulate complex phenomena \cite{kaufmann1992supercomputing}. The second generation of simulation technology heavily depends on formulae and hypotheses summarised by humans based on theories and real-world observations. However, mathematical models are not efficient enough to represent all the laws in the world. 
	
	Driven by the success of the fourth paradigm, the third generation of simulation emerged. In the third generation of simulation, advanced algorithms make machine learning models able to learn data distribution. This progression reduces the necessity for researchers to summarise causal relationships in the real world \cite{mayer2013big}. In challenging data collection scenarios, the third generation of simulation can generate data for research purposes. In the realm of the Internet of things (IoT), a knowledge-enhanced generative adversarial model effectively produced network traffic data, addressing the significant challenge of acquiring large-scale IoT traffic data \cite{hui2022knowledge}. Similar methods were utilised in personal activity trajectories data \cite{yuan2022activity} and other daily life research \cite{yuan2023learning}.
	
	Although these methodologies can recognize patterns from data, its prediction or classification performance often suffers when encountering individual-level samples not consistent with the historical data distribution, i.e. the uncertainty of dynamic distribution in problems of organised complexity.
	
	Therefore, we have to focus on the \textit{next-generation simulation} (NGS), scrutinising a system through more authentic individual simulation to gather feedback, and it is a possible solution to problems of organised complexity. According to the conclusions of Lorenz \cite{lorenz1963deterministic} and May \cite{may1974biological}, the intrinsic nature of chaotic dynamical systems suggests an inherent inability to fully comprehend the intricate correlations among individuals and the latent patterns within chaos. Consequently, the recourse to understanding chaotic system states primarily lies in simulation, aside from the limited scope of observing real-world systems. Moreover, the more authentic one simulation is, the more informative its results will be. If we can better simulate sophisticated human beings, problems concerning high-order chaos will be better analysed. Advanced to conventional agent-based modelling simulation (ABMS), the methodologies to realise the NGS can be in the form of a higher degree of paradigms integration to simulate individual-level behaviours of dynamic distributions.
	
	We propose \textit{sophisticated behaviour simulation} (SBS) as a technique of the NGS in complex human systems where sophistication like mental factors and learning mechanisms have to be considered. SBS draws inspiration from the term ``behavioural economics'' \cite{angner2007behavioral}. According to the terming of ``behavioural economics'', the modifier behavioural has historical significance, originating from behavioural decision research \cite{payne1992behavioral}. Behavioural decision research, in essence, is dedicated to studying the intricacies of how individuals make decisions. Historically and theoretically, it focuses on comparing actual decision-making processes with the principles of rationality \cite{dawes1998behavioral}. It is from this foundation that behavioural economics emerged as a distinct field. Behavioural economics focuses on the study of irrational human behaviour, recognising that individuals often deviate from purely rational decision-making processes. The term ``behavioural'' highlights the departure from traditional economic models that assume strict rationality. By incorporating ``behavioural'' into the term, we aim to create simulations that authentically represent the complexities, biases, and nuances inherent in human decision-making to scrutinise complex systems of high-order chaos. 
	
	The agents in existent ABMS generate their behaviours based on historical data following a fixed pattern. Human behaves according to his observation, experience, learning, and culture, and these factors account for the sophistication. Therefore, its generation have to be more interpretable and follows dynamic patterns to be consistent with human, a typical characteristic of organised complexity. Contrary to agents in ABMS, the agent in SBS should generate its behaviours based on its \textit{knowledge architecture}, which can be updated through a \textit{learning mechanism}.
	
	\section{Realisation of Sophisticated Behavioural Simulation}
	
	\begin{figure}
		\centering
		\includegraphics[width=1.0\columnwidth]{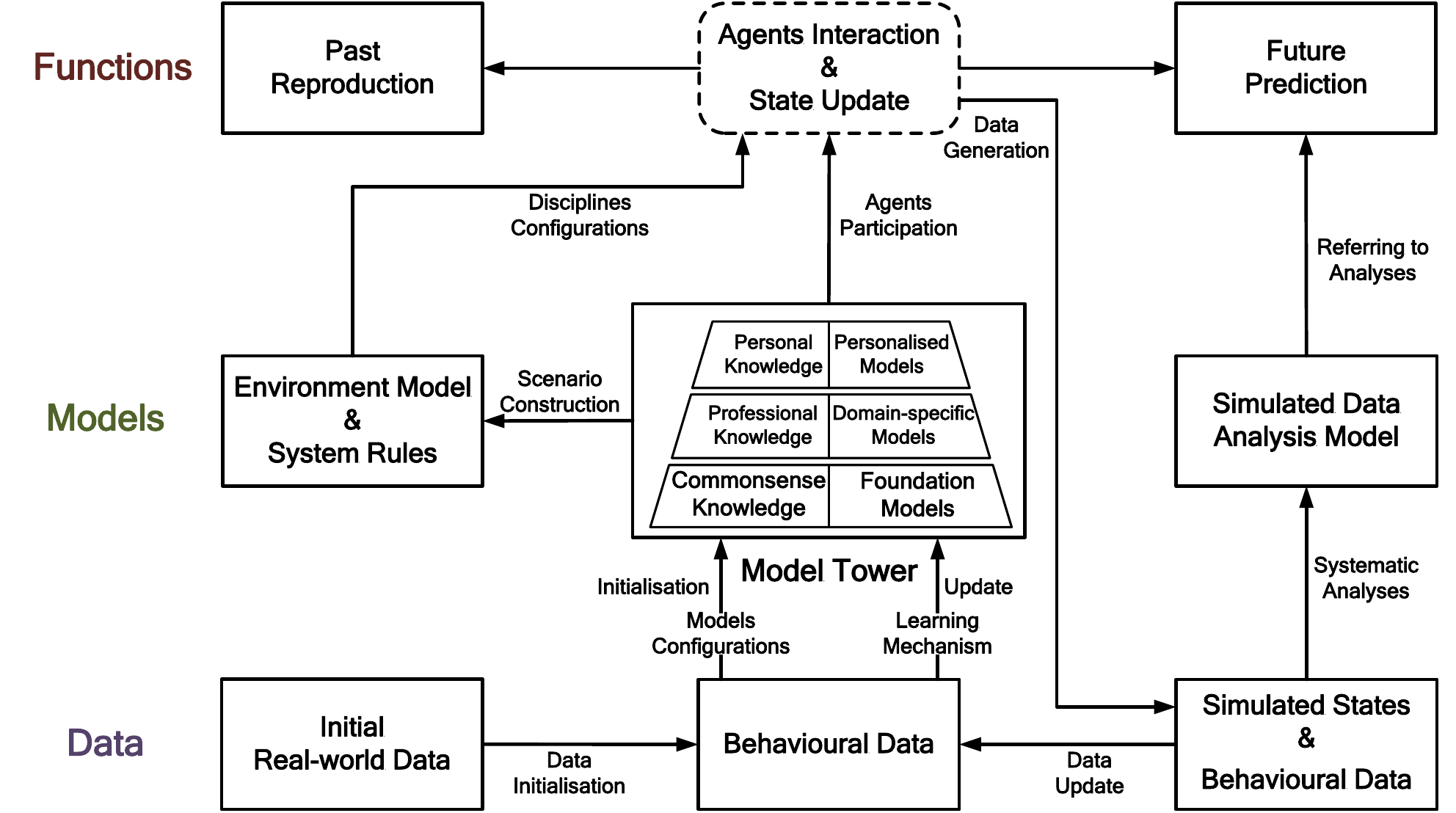}
		\caption{The framework of SBS.}
		\label{fig:SF}
	\end{figure}
	
	To realise the so-called \textit{knowledge architecture} and \textit{learning mechanism}, we propose \textit{the sophisticated agent}. Within the sophisticated agent, different levels of knowledge are included by models of different paradigms. Through the fusion of models, the knowledge can form a hierarchical knowledge architecture. With the sophisticated agent as the core, importing learning mechanism, we design the architecture of SBS as illustrated in Figure \ref{fig:SF}. The framework of SBS is separated into three layers: 
	\begin{itemize}
		\item \textbf{Functions} 
		\begin{itemize}
			\item The process of simulation is the basic function in SBS.
			\item Simulation results are further utilised for past events production and future prediction.
		\end{itemize}
		\item \textbf{Models} 
		\begin{itemize}
			\item The core is our proposed \textit{model tower} which consists of three floors. After the model tower empowers instances of the sophisticated agents, these agents become a part of the simulation scenario and participate in interactions within the system.
			\item Environment analysis model and system update rules confirm the essence of the system which is the disciplines of agents' interactions and states update.
			\item The simulated data analysis model is designed to analyse systemic data so that its output can be the reference for future prediction. 
		\end{itemize}
		\item \textbf{Data} 
		\begin{itemize}
			\item Initial real-world data are collected to initialise the model tower as the seed of SBS.
			\item Simulated states/behavioural data are generated in the system states update and interactions among agents after every tick. Simultaneously, a learning mechanism affects models in the model tower to update agents' hierarchical knowledge systems.
			\item Initial data and simulated data constitute all behavioural data in SBS.
		\end{itemize}
	\end{itemize}
	
	\subsection{Hierarchical Knowledge Architecture: Model Tower}
		
		The sophisticated agent adopts an architecture we name \textit{model tower}, comprising self-organised multi-floor models. We aim to generate authentic behaviours by initially feeding current system states and other input streams into the bottom floor in a specific format. As the data ascends through each floor, it undergoes bottom-up refinements, culminating in the generation of finely nuanced individual-level behaviours. The model tower is considered a research tool of paradigms integration where deep learning model is the basis, and upper floors could be the second, third, and fourth paradigm models. This is what we call a higher degree of integration in SBS than in existent ABMS methods. The model tower builds a sophisticated behaviour generation framework and hierarchical knowledge architecture containing commonsense knowledge, professional knowledge, and personal knowledge.
		
		\subsubsection{Commonsense Knowledge}
		
		The bottom floor conducts general reasoning based on commonsense knowledge and establishes a preliminary direction for subsequent refinements. As a foundation model, it typically serves as the base for this purpose. Among existing research \cite{park2023generative, gao2023s, wang2023voyager, li2023camel}, LLM stands out as a predominant choice due to its remarkable reasoning capabilities and extensive inherent commonsense knowledge. Through prompting engineering, the LLM agents are able to apply their knowledge and think as human. To make the most use of commonsense knowledge, there are plenty of tricks behind the design of prompts like templates and cognitive architectures \cite{sumers2023cognitive}, and it is closely related to the application scenarios. As the rapid upgrade of LLMs and other forms of foundation models, they will be doomed to possess more and more powerful commonsense reasoning ability. What the basis of the model tower is required to do is align prior knowledge with human in specific scenarios through well-designed prompts.
		
		\subsubsection{Professional Knowledge}
		
			\begin{figure}[!t]
				\centering
				\includegraphics[width=0.8\columnwidth]{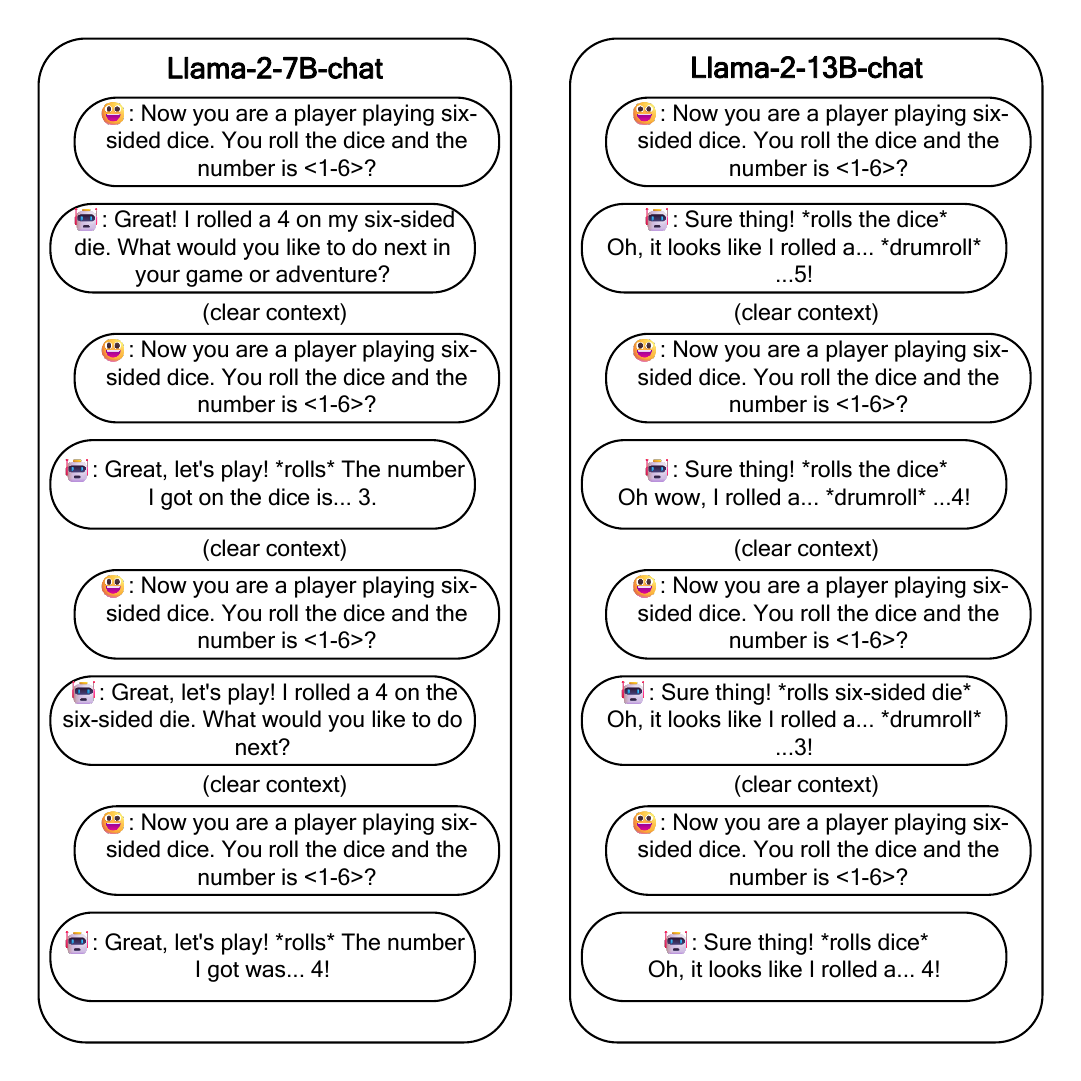}
				\caption{The dice rolling experiments was conducted on LLaMA-2-7B-chat and LLaMA-2-13B-chat.}
				\label{fig:DR}
			\end{figure}
			
			\textit{1) Internal Specialised Generative Module: }As depicted in Figure \ref{fig:DR}, the sequence of dice rolls generated by LLM is intuitively peculiar. To ensure the independence of each roll, we clear the context for every roll request. Despite these rolls being mutually independent, the number ``4'' emerges as the most frequent outcome. Notably, with either LLaMA-2-7b-chat or LLaMA-2-13b-chat \cite{touvron2023llama}, the numbers ``1'', ``2'', and ``6'' never appear. Even when considering that this sequence holds a probability the same as the sequence 123456123456... (which stands at $\frac{1}{6^n}$ when using a standard dice), the observed likelihood remains peculiar.
			
			In the dice rolling problem, if the probabilities assigned to the six sides appearing in the candidate tokens differ, the model struggles to accurately simulate real dice rolling, which typically follows an equal probabilities model.
			
			In comparison with the hallucination of LLM, the generated behaviour isn't logically incorrect but incorrect in probability. While obtaining a sequence where ``4'' consistently appears in a finite number of dice rolls is probable, the likelihood of it being a standard dice is notably low.
			
			The specialised generative models within the model tower will effectively address this issue. Introducing an additional floor above the foundational model floor facilitates a two-floor architecture, separating reasoning from generation into distinct processes. This significantly mitigates the dice rolling problem. In the dice rolling scenario, we first prompt the LLM to generate ``\{`action': `dice rolling'\}''. Then, the specific generative module processes the request, calling a generative model whose distribution is the same as a standard six-sided dice to generate the final action ``\{`dice rolling': 4\}''. Theoretically, the result sequence generated in this pattern can be far more authentic than the basic LLM agent. This pattern allows LLM to perform commonsense reasoning, while invoking next-floor domain-specific generative models, for reliable behaviour generation supplemented by professional knowledge in the form of distributions.
			
			\begin{flushleft}
				\textit{2) External Application Programming Interface: }The division of reasoning and generation across floors within the aforementioned model tower architecture is similar to the application programming interfaces (APIs) calls observed in online language assistant agents such as Copilot and retrieval-augmented generation (RAG) of LLMs \cite{lewis2020retrieval}. Numerous efforts, as highlighted by Patil et al., \cite{patil2023gorilla} have focused on enhancing the comprehensive abilities of LLM agents. Therefore, beyond the specialised generative module, leveraging various external APIs, such as calculators, databases, and others holds promise in mitigating hallucinations and improving reasoning, which is also a good choice when the time and hardware costs are acceptable.
			\end{flushleft}
			
		In summary, the internal specialised generative module and external APIs collectively form the second floor within the model tower, relieving the first floor (foundation model) from professional inference. As a result, most task-specific professional knowledge is obtained in the second floor through agents calling APIs. Plenty of work about professional LLM agent has verified the feasibility and necessity of the import of professional knowledge \cite{wang2024large, chu2024professional, zhu2024knowagent}.
		
		\subsubsection{Personal Knowledge}
		
		Dramaturgy proposed by Erving Goffman \cite{goffman1959presentation} states that everyone's social behaviours heavily rely on societal norms. Besides external factors, various internal elements distinctly influence one's behaviours, including character, experience, gender, occupation, mood, etc. Therefore, to achieve the sophisticated agent, a comprehensive profile of an agent containing the aforementioned external and internal information is required to differentiate agents driven by a two-floor model tower. Above aforementioned impersonal knowledge floors, a floor provides group and individual bias is required. We call this floor personalised models floor. Rather than relying solely on textual profiles \cite{gao2023s}, the more sustainable resolution is adding a personality model upper the two-floor model. This addresses the issue between the capacity of the LLM context and the intricate nature of personality. In other words, the personalised models can be realised in the form of rules, which belongs to the second paradigm. A better resolution is using personality models, which belongs to the fourth paradigm.
		
	\subsection{Behavioural Learning Mechanism: Reinforcement Learning}
	
	 As the simulation proceeds, a behavioural learning mechanism is utilised to update agents' hierarchical knowledge systems. The behavioural learning mechanism empowers agents with flexibility, enabling them to adjust their knowledge system and behaviours in response to feedback of interactions and environment. This mechanism enables agents to learn from mistakes, self-improve, and continuously optimise their decision-making and behavioural patterns. Through continuously accumulating experiences and refining their performance, agents enhance their efficiency and strategy in decision making. The learning mechanism is crucial for agents, and the reinforcement learning (RL) method is widely selected as the learning mechanism for agents owing to its high flexibility, self-learning, experience accumulation, etc. Post the agent's generation of pertinent prompts derived from its current observations, CoT \cite{wang2022self, wei2022chain} combining with RL mechanism engages in intricate reasoning to determine the optimal behavioural scenarios. Besides, the behavioural learning mechanism should be implemented on the update of upper-floor models in the model tower in order to realise fine-grained behavioural learning.
		
	\begin{figure*}[!t]
		\centering
		\subfloat[Causality Analysis]{
			\includegraphics[width=0.45\columnwidth]{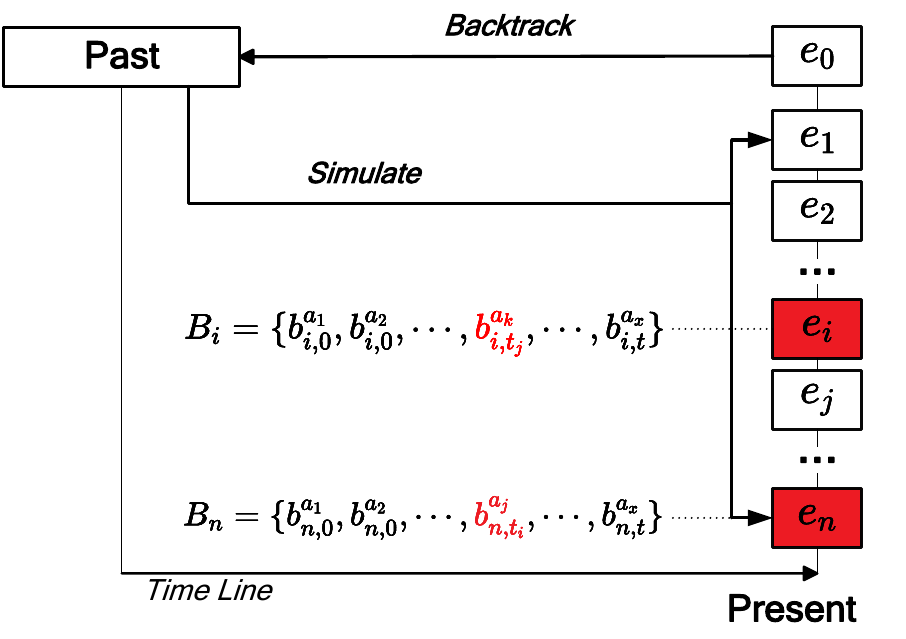}
			\label{fig:SDA1}
		}
		\subfloat[Rule-based Analysis]{
			\includegraphics[width=0.45\columnwidth]{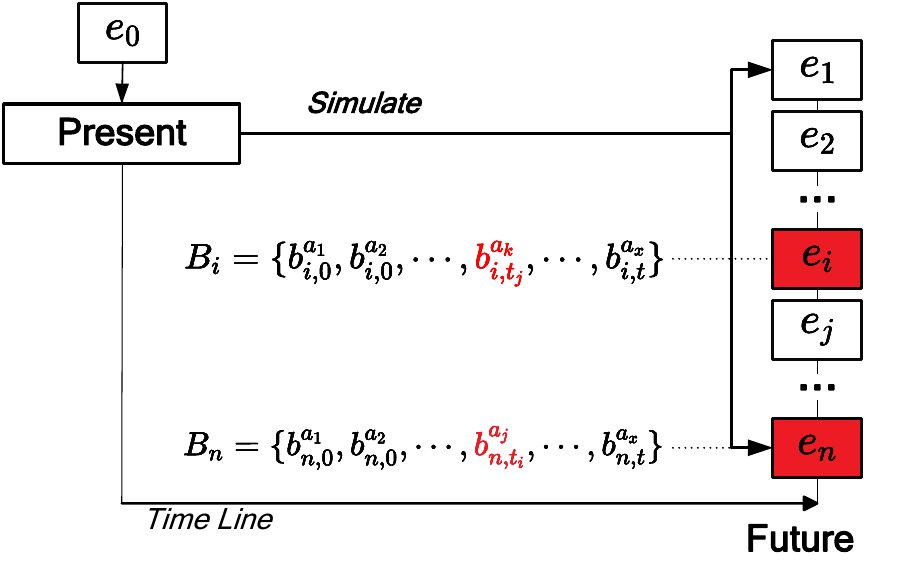}
			\label{fig:SDA2}
		}
		\qquad
		\subfloat[Deep Learning Analysis]{
			\includegraphics[width=0.776\columnwidth]{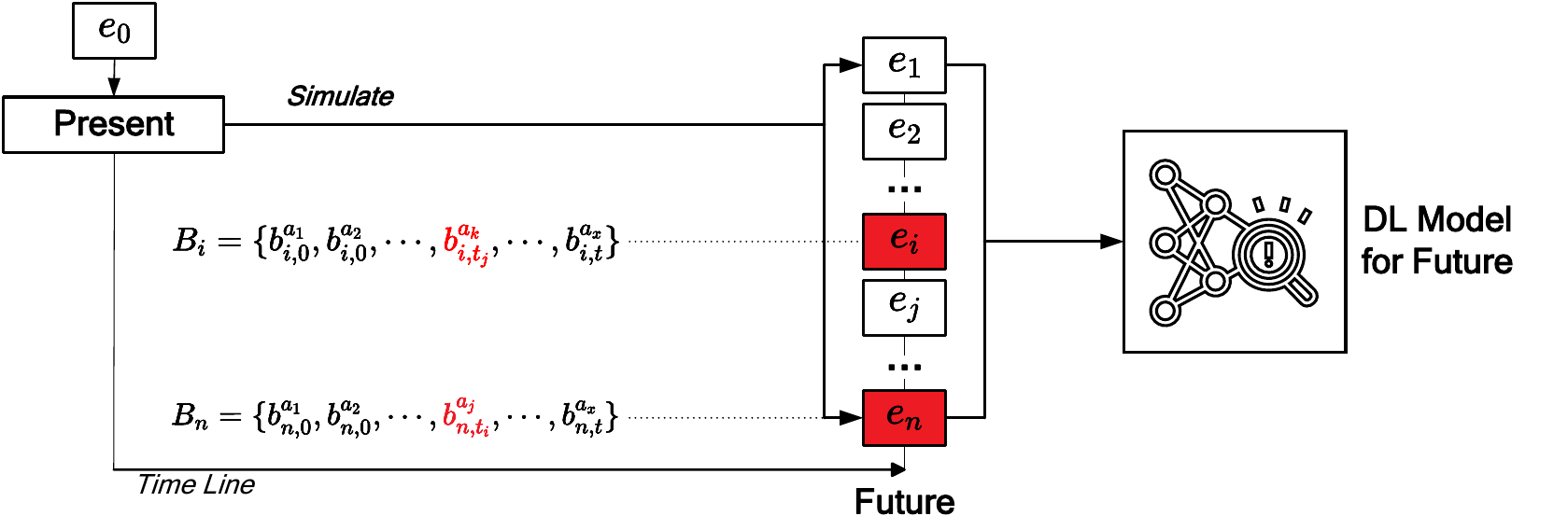}
			\label{fig:SDA3}
		}
		\caption{(a) The illustration of causality analysis. In this example, $e_i, e_n$ have similar special states to $e_0$, and the behaviour $b_{i,t_j}^{a_k}, b_{n,t_i}^{a_j}$ of the simulated data $B_i, B_n$ is suspected by manual analysis. (b) The illustration of rule-based analysis. In this example, $e_i, e_n$ have the special states unexpectedly disobeying some rules, and through manual analysis of simulated data $B_i, B_n$ behaviour $b_{i,t_j}^{a_k}, b_{n,t_i}^{a_j}$ is suspected. (c) The illustration of deep learning analysis. In this example, $e_i, \cdots, e_n$ are simulated futures. By training a deep learning model based on simulated data, it will be able to detect anomalies like $b_{i,t_j}^{a_k}, b_{n,t_i}^{a_j}$ and further predict $e_i, e_n$ before $e_i, e_n$ happen.}
	\end{figure*}
		
	\section{The Utilisation of Next-Generation Simulation}
		
	To deal with the problems of organised complexity, the NGS needs to scrutinise issues about system states. Given that accurate prediction of states within chaotic systems is impossible, it implies that simulations may not perfectly correspond with the actual evolution of these systems. As a result, it becomes crucial to ensure the generation of diverse initial states that yield varied ``world lines'' for meaningful analysis. The more simulations are conducted, the higher the likelihood of capturing the possible future trajectory of a system will be. By leveraging the interpretability inherent in logs of agents' behaviours and thoughts generated by foundational models such as LLMs, we primarily present some analysis patterns for two functions in SBS.
	
	\subsection{Past Reproduction}
		
		SBS makes past reproduction possible. Through the details of simulated events, causality analyses on abnormal behaviours and events can be conducted. This has the potential to be utilised in scenarios such as security events investigation and review in complex human systems. When some unexpected states or events arise in a real-world system (represented by $e_0$), simulating the system from a particular past moment becomes a viable approach. As is shown in Figure \ref{fig:SDA1}, conducting numerous simulations yields a spectrum of diverse states or events, denoted as $E=\{e_1, \cdots, e_n\}$, occurring when the simulated system progresses to the same point as $e_0$. Among these states or events, we filter those resembling $e_0$, forming the set $E_\tau=\{e_i, e_j, \cdots, e_k\}$. Subsequently, we can investigate the individual behaviours $B_i=\{b_{i, 0}^{a_1}, b_{i, 0}^{a_2}, \cdots, b_{i, T}^{a_x}\}$ ($b_{i, T}^{a_m}$ represents the behaviour of agent $a_m$ at moment $T$ before $e_i$) leading to the $e_i \in E_\tau$. Employing manually checking the common features in $B_\tau=\{B_i, B_j, \cdots, B_k\}$, meaningful conclusions regarding the causality hardly noticed in complex systems underlying the appearance of $e_0$ may be drawn.
		
	\subsection{Behavioural Rehearsing}
		
		SBS can be instrumental in predicting future risks, utilised in scenarios such as safety events warning and avoiding. We name it \textit{behavioural rehearsing}, discovering potential problems through numerous times of behavioural simulations in advance. Similar to the methodology employed in causality analysis, the simulation process initiates from the present moment. As is shown in Figure \ref{fig:SDA2}, as the simulation progresses, diverse states or events $E=\{e_1, \cdots, e_n\}$ are observed at a specified future moment $t$. In this scenario, predefined rules serve as benchmarks for expectations at moment $t$. These rules facilitate the identification of unexpected future occurrences, denoted as $E_\beta$. If $E_\beta=\emptyset$, indicating no unexpected outcomes, it signifies that the future aligns with expectations. However, if $E_\beta \neq \emptyset$, it indicates that the simulation predicts deviations from expectations due to certain specific and notable behaviours. In such cases, manual intervention becomes imperative to prevent these behaviours.
		
		Besides human-assist analyses, a completely autonomous approach is feasible. As is shown in Figure \ref{fig:SDA3}, regarding a system as a graph where nodes are agents, and edges are interaction behaviours between agents, each simulation generates a series of graphs. By aggregating these graphs from multiple simulations, a comprehensive dataset emerges. This dataset becomes the basis for training deep learning models. This process is revolutionary as it enables training deep learning models even when no real-world data is available. Such a methodology proves indispensable, especially in scenarios where acquiring data from high-order chaotic systems is challenging or unfeasible \cite{wang2024large}, just as resorting to synthetic images supplement datasets to avoid expensive dataset annotations \cite{8099724}.
	
	\section{Conclusion and Outlook}
	
		Among the three types of scientific problems, problems of organised complexity pose the greatest challenge. To resolve them, great efforts have been made in developing the three cornerstones of scientific research: data, algorithms, and computational power. The four paradigms have sequentially emerged to tackle the challenges in increasingly complex scientific research fields. As representative methods of the third and fourth paradigms, simulation and machine learning technologies have become the solution to many problems of organised complexity. Considering the uncertainty of unresolved problems, we find that high-order chaos renders relying solely on one paradigm ineffective. Therefore, drawing on the concept of paradigms integration, we explore a form of the NGS and propose SBS as a methodology. SBS integrates the second, third, and fourth paradigms to achieve more authentic simulation in such systems. To initiate and enhance SBS, we propose the sophisticated agent driven by the model tower, and discuss how to utilise the outputs of the NGS.
		
		However, it is crucial to acknowledge the limitations and challenges inherent in the NGS. The intrinsic uncertainty within organised complex systems precludes precise future prediction or past reproduction, similar to the concept of Laplace's demon \cite{laplace2012philosophical}, a concept deemed unrealistic. The NGS serves as a valuable reference rather than an absolute predictor, while deep learning models excel at specific prediction and classification tasks. Challenges persist, including determining the required width of the context window, achieving personalised modelling without text profiles, implementing experiential learning throughout the model tower, and estimating the resource demands of large-scale SBS.
		
		We are looking forward to the application, and other methodologies of the NGS. Their advancement, coupled with the benefits derived from the development of other paradigms, positions the NGS as a promising and risk-avoiding methodology. This ensures it indispensable in cost-sensitive social science scenarios such as policy piloting, financial strategy, and systematic risk prediction, where real-world experiments may not be feasible. 
	
	\bibliography{BehaviouralSimulation}

\end{document}